\documentclass[lettersize,journal]{IEEEtran}
\usepackage{amsmath,amsfonts}
\usepackage{algorithmic}
\usepackage{algorithm}
\usepackage{array}
\usepackage[caption=false,font=normalsize,labelfont=sf,textfont=sf]{subfig}
\usepackage{textcomp}
\usepackage{stfloats}
\usepackage{url}
\usepackage{verbatim}
\usepackage{graphicx}
\usepackage{cite}
\usepackage{hyperref}
\usepackage{url}

\newtheorem{theorem}{Theorem}[section]

\newtheorem{lemma}[theorem]{Lemma}

\newtheorem{assumption}[theorem]{Assumption}
\newtheorem{remark}[theorem]{Remark}
\newtheorem{example}[theorem]{Example}
\usepackage{graphicx} 

\title{Communication and Energy Efficient Federated Learning using Zero-Order Optimization Technique}
\author{Elissa Mhanna and Mohamad Assaad\\ Laboratory of Signals and Systems (L2S), CentraleSupelec, University of Paris-Saclay, France\\Emails:\{Elissa.Mhanna, Mohamad.Assaad\}@centralesupelec.fr }
\date{August 2024}

\begin{document}

\maketitle

\begin{abstract}
Federated learning (FL) is a popular machine learning technique that enables multiple users to collaboratively train a model while maintaining the user data privacy.   A significant challenge in FL is the communication bottleneck in the upload direction, and thus the corresponding energy consumption of the devices, attributed to the increasing size of the model/gradient. In this paper, we address this issue by proposing a zero-order (ZO) optimization method that requires the upload of a quantized single scalar per iteration by each device instead of the whole gradient vector. We prove its theoretical convergence and find an upper bound on its convergence rate in the non-convex setting, and we discuss its implementation in practical scenarios. Our FL method and the corresponding convergence analysis take into account the impact of quantization and packet dropping due to wireless errors. We show also the superiority of our method, in terms of communication overhead and energy consumption, as compared to standard gradient-based FL methods. 
\end{abstract}


\section{Introduction}
\label{intro}

Machine Learning and mainly Neural Network (NN) schemes are gaining popularity in several areas and applications. With the massive amount of data generated by wireless devices, distributed learning techniques are attracting increasing interest in both sectors of academia and industry \cite{FLGoogle}. 
Among the innovative techniques, Federated Learning (FL) \cite{FL1} is an interesting approach that allows multiple edge devices to train collectively a given model while ensuring the data privacy of the devices. 

In FL, the edge devices use their own data to locally update the model using the gradient of the loss function and send back the gradients or the locally updated model to the server. The latter aggregates the received models or gradients and sends back the aggregated model to the devices, and the process repeats. Improvements in the efficacy of FL have been made using first-order \cite{FL1, FL2,FL3} and second-order \cite{FLS, FLS2} methods. However, these techniques may require high communication and computation resources, especially when used over wireless systems \cite{FLchallenges}. For instance, with the rapid development of wireless systems and mobile devices, coupled with the vast amounts of data generated in these networks, there is growing interest in applying FL within these wireless systems. However, the use of standard FL methods in this case implies some challenges. 
In fact, since the model/gradients have a high dimension of  $d$, $d$  values must be communicated by each device to the server in each round of the FL. This represents a fundamental communication bottleneck in FL and results in a high energy consumption of mobile devices. It is worth mentioning that the uplink of wireless networks, especially due to the limited power of the devices, may afford the huge amount of communication required by FL. 
To deal with the communication bottleneck, some existing work has considered that devices perform local multiple gradient descent steps before sending their gradients or model updates to the server \cite{localSGD}. Saving of Communication resources can also be done by partial device participation at every iteration \cite{pp}\cite{FL1}. 
Lossy compression of the gradients before uploading it to the server is also considered in \cite{CQ1, CQ2, CQ3}, where stochastic unbiased quantization approaches are used. Quantization of gradient differences between two iterations is considered in current and previous iterations
\cite{CQ4}, allowing thus the update to incorporate new information. Sparsification of this difference, in the sense that components that are not large enough will not be transmitted,  is also considered in \cite{S1} to reduce the amount of information sent in the uplink.

In this paper, we approach this problem in a novel way: instead of using the gradient method, we adopt a zero-order approach and develop a zero-order training method that requires to send one scalar only instead of a long gradient vector. This reduces highly the number of variables communicated over the wireless links and results in a reduction in terms of energy consumption. The impact of quantization, used in digital systems nowadays, and the packet dropping due to wireless errors are included in our framework. Zero-order (ZO) methods are a subfield of optimization where it is assumed that the gradient is not available or cannot be computed. In ZO optimization, the gradient is estimated using function values queried at a certain number of points \cite{ref2, ref2-ref}.

In \cite{allerton}, ZO is studied where it is assumed that information is transmitted using analog communication, while in practical systems digital communications is used. In this paper, we extend hence the use of ZO method to the digital context. The challenge here lies in the fact that quantization is used which results in an error that may compromise the convergence to the desired result. In this work, we propose an algorithm where also only a quantized scalar is sent in both uplink and downlink phases of FL, which  significantly enhances the efficiency of our method. In addition, since information is transmitted over the wireless links, errors due to the channel fading can occur. In this paper, we consider this issue as well and include its  impact on the convergence analysis of our proposed method.  
In this paper, we deal with a set of challenges. First of all, unlike most of existing work that studies the performance of FL over wireless links using gradient method and in a convex setting, we consider a more realistic setting in which the objective function is not convex. In fact, it is widely known that loss functions in FL and NN are in general not convex. Addressing nonconvexity in FL is challenging. Our ZO method must hence handle nonconvexity, quantization, noise, and wireless channel stochasticity, which can  prevent an iterative method from convergence. Unlike convex cases, nonconvex optimization does not allow easy quantification of optimization progress. Furthermore, since our method is based on a ZO technique, it leads to a biased gradient estimate. Verifying convergence becomes intricate due to the bias term, especially in the nonconvex setting. 
In this work, we overcome these difficulties and propose a new communication and energy-efficient algorithm in the nonconvex setting. By carefully examining our method's setting, we are able to tailor a biased gradient estimate from scalar feedback sent in both directions and analyze it intricately using probabilistic tools. We study scrupulously its expectation, noting the stochastic processes influencing these scalars, and give equal effort to bounding its norm squared. We then prove rigorously the convergence of our method utilizing tools from stochastic approximation methods, emphasizing on the evolution of the sum of exact gradient and its interplay with the step sizes. Expanding on the step sizes' form, we provide the convergence rate, which competes with the standard gradient method while saving a lot of execution time and communication overhead.  
  Finally, we provide a thorough analysis of the energy consumption reduction achieved by our method. 

The advantages of our approach are three fold. It allows countering the communication and energy consumption bottlenecks of FL, by limiting the exchange between the devices and the server to scalar-valued updates, which allows saving up to a factor of $O(d)$ in both transmission directions) in comparison to standard methods. Furthermore, unlike standard methods that require computational capabilities to compute the gradients at each device, e.g. using backward propagation, our approach does not need to find the gradient since each device computes a numerical value of the loss function at each iteration using forward propagation. The above two points result in an energy consumption reduction. Finally, our method is suitable to the cases where the gradient is complicated to compute, which arises in several examples in practice, e.g.  in hyperparameter tuning where there is no closed form expression of the loss function with respect to the hyperparameters \cite{ZO-Bay}.

\section{System Model and proposed algorithm}
\subsection{System Model}
We consider a federated learning setting where $N$ edge devices collectively train a model over a wireless network by coordinating with a central server over a wireless network. Each device has its own private data, and the exchanges between the devices and the central server are limited to the model parameters (i.e., optimization parameters).  Let $\mathcal{N}=\{1,...,N\}$ be the set of devices and $\theta\in\mathbb{R}^d$ denote the global model to be found. Let $F_i:\mathbb{R}^d\rightarrow\mathbb{R}$ be the loss function associated with the local data stored on device $i$, $\forall i\in\mathcal{N}$. The  objective is to minimize the function $F:\mathbb{R}^d\rightarrow\mathbb{R}$ that is defined by the loss functions of the devices as follows,  
\begin{equation}\label{F_i}
	\min_{\theta\in\mathbb{R}^d} F(\theta):= \sum_{i=1}^{N} F_i(\theta)\;\;\;\text{with}\;\;\;	F_i(\theta)=\mathbb{E}_{\xi_i\sim D_i} f_i(\theta,\xi_i).
\end{equation}
$\xi_i$ is an i.i.d. ergodic stochastic process following a local distribution $D_i$, that models local data distribution. We consider that functions $F$, $F_i$, and $f_i$ are nonconvex. In standard FL methods, each device updates the model locally by computing the gradient of the loss function $F_i$ and then uploading their local gradients or models to the server. This requires computation and high communication overhead since the gradient is a long vector of size $d$. For instance, $d$ is in the order of hundreds of thousands of parameters in practice. In this paper, we avoid the computation and exchange of the gradients by adopting a ZO approach.  The devices query their model only once per iteration and obtain a scalar value from this query that is sent back to the server. In addition, we consider that each query is quantized before being sent.\\ 
Furthermore, the quantized scalars exchanged over the wireless links are subject to fading and are not necessarily received correctly. 
We consider the server is able to receive and correctly decode a packet from user $i\in\mathcal{N}$ with probability $0<p\leq 1$. Otherwise, the packet is considered erroneous and is dropped. We further assume that the users' channels are independent of each other. At every time or iteration $k$ of the FL, we consider that the users with the correctly decoded packets belong to the set $\mathcal{S}_k$, i.e.,
	\begin{equation}
		\mathbb{P}(i\in\mathcal{S}_k)=p \;\;\;\text{and}\;\;\; \mathbb{P}(i\notin\mathcal{S}_k)=1-p.
	\end{equation}
\subsection{Algorithm}
This section provides a simple digital zero-order federated learning (DZOFL) method. 

The algorithm is described as follows. At each iteration, every user $i\in\mathcal{N}$ computes two queries of its loss function using its local data and then computes their difference, i.e., $\Delta f_{i,k} = f_i\big(\theta_k + \gamma_k\Phi_k, \xi_{i,k}\big)-f_i\Big(\theta_k - \gamma_k\Phi_k, \xi_{i,k}\big)$,  where $\Phi_k$ is a perturbation direction generated randomly and pre-stored in the devices, $\gamma_k$ is the step size that will be specified later in the paper, and  $\theta_k$ is the model at iteration $k$.  Then, user $i$ applies a quantizer operator $Q(\cdot)$ on this difference of queries and sends the quantized value/scalar to the server, denoted as $Q(\Delta f_{i,k})$. Once these quantized values are sent to the server, the latter decodes the quantized signals and combines them (from all devices) if they are received correctly; that is, the server receives
\begin{equation}\label{reception}
	\Delta f_k = 
	\begin{cases}
		\frac{N}{|\mathcal{S}_k|}\underset{i\in\mathcal{S}_k}{\sum} Q(\Delta f_{i,k}), &\text{if } |\mathcal{S}_k|\neq 0,\\
		0, &\text{if } |\mathcal{S}_k|= 0.
	\end{cases}
\end{equation}
The packets that are not received correctly are dropped. The server then quantizes this aggregated scalar, denoted as $Q(\Delta f_k)$, and broadcasts it to all users. The model is then updated by each user via $\theta_{k+1} = \theta_k - \alpha_k g_k$, where $g_k$ is given in (\ref{2p_grdt_est}), and $\alpha_k$ is a step size that will be specified in the next section. 
\begin{equation}\label{2p_grdt_est}
	\begin{split}
		g_k = & \Phi_k \times Q(\Delta f_k)\\
		= &\Phi_k Q\Bigg(\frac{N}{|\mathcal{S}_k|} \sum_{i\in\mathcal{S}_k} Q (\Delta f_{i,k})\Bigg)
	\end{split}
\end{equation}

In standard methods, the gradient descent step is applied at the server and then sent back to the devices. However, in our proposed technique, the server simply sends the scalar $\Delta f_k$, and each device computes locally $g_k$. 
It is worth noticing that whenever the set $\mathcal{S}_k$ is empty, or $|\mathcal{S}_k|= 0$, all users' packets were found erroneous. The server thus cannot estimate the gradient and must keep the same model $\theta_{k+1}=\theta_k$ for the next round. 
$\Phi$ - the perturbation direction generated randomly - can be generated by the server and communicated to the devices before the training. $\Phi$ can hence be seen as a universal perturbation implemented offline in the devices and used in all training. Another way to generate $\Phi$ is to use Shift Registers (e.g., similar to the Pseudo Noise sequence in CDMA systems). The devices will use the same polynomial generator, already predefined for all devices, and the server simply sends a short vector containing the initial values of the registers at the beginning of the training process.  
We present our proposed method in Algorithm \ref{alg:example_2p}.

\begin{remark}
   One can see that each device sends only a quantized scalar in the uplink instead of a long gradient vector, as in standard FL. This results in huge savings in communication overhead. In addition, the devices do not need to compute the gradient which saves energy and computation resources.  
\end{remark}
 
\begin{algorithm}[h]
	\caption{The DZOFL Algorithm}
	\label{alg:example_2p}
	{\bfseries Input:} Initial model $\theta_0\in\mathbb{R}^d$, the initial step-sizes $\alpha_0$ and $\gamma_{0}$, the initial perturbation vector $\Phi_0$ 
	\begin{algorithmic}[1]
		\FOR{$k=0, \ldots, K$}
		\STATE Let $\theta_k$ be the model at iteration $k$. Every user $i$ queries the loss function for $\theta_k + \gamma_k\Phi_k$ and $\theta_k - \gamma_k\Phi_k$  with its local data and computes the difference of the queries $\Delta f_{i,k}=f_i\big(\theta_k + \gamma_k\Phi_k, \xi_{i,k}\big)-f_i\Big(\theta_k - \gamma_k\Phi_k, \xi_{i,k}\big)$.
		\STATE Every user $i$ quantizes this difference of queries (scalar), $Q(\Delta f_{i,k})$, and sends it back to the server
		\STATE The server aggregates the received quantized queries\\ $\frac{N}{|\mathcal{S}_k|}\underset{i\in\mathcal{S}_k}{\sum} Q\Big(\Delta f_{i,k}\Big)$
        \STATE The server quantizes this aggregated scalar, $Q\Big(\frac{N}{|\mathcal{S}_k|} \sum_{i\in\mathcal{S}_k} Q (\Delta f_{i,k})\Big)$, and sends it back to the devices\\
		\STATE The model is updated using $\theta_{k+1} = \theta_k - \alpha_k g_k$, where $g_k$ is given  in (\ref{2p_grdt_est})\\
		\ENDFOR 
	\end{algorithmic}
\end{algorithm}

\section{Energy and communication overhead reduction}
In this section, we discuss the energy consumption and communication overhead of our proposed algorithm. We also compare it with standard FL to explain the superiority of our proposed method. The values provided here are only examples of potential values that can be obtained in the implementation of FL methods. The exact numerical results are provided in Section \ref{num}. The goal of this section is to explain why our method can provide better performance in terms of communication overhead, convergence time, and energy efficiency compared to the standard FL method.

\subsection{Communication overhead}
Let $N$ be the number of devices. $T$ is the number of iterations needed to converge. Per iteration, in our method, each device sends one quantized scalar of $M$ bits. Therefore, the total number of upload bits is $TNM$ bits. 

In standard FL, e.g., \cite{FL1}, the number of iterations is $T'$. Per iteration, there are $N$ devices sending each a long vector of $d$ quantized scalars, that is $Md$. The total amount of uploaded information is $T' N M d$. In standard FL, gradient descent with FO information is used, which is usually faster than ZO methods, that is, $T'<T$. To illustrate the potential gain of our method, let us consider that $T'$ is around $100$ to $1000$ iterations, which are reasonable values according to existing literature in this area. Let us consider also that $T$ is  $10$ times higher. On the other hand, the length of the gradient vector $d$ is huge. For example, in many classification problems and other ML examples, the number of parameters varies between a few hundred thousand to a few million, depending on the NN architecture. To illustrate the gain of our method, let us say $d=4\times 10^5$. For $M=16$ bits, one can see that the total number of bits uploaded in the standard method is  $64\times 10^7$ bits per device for $T'=100$ iterations,  while in our method, is $16\times 10^3$, i.e., $40000$ times less than standard FL (for $T=10000$). One can see that even if an efficient compression technique is used in standard FL, the amount of information will be much higher than our method. It is worth mentioning that usually, $d$ could be much higher than $4\times 10^5$. For instance, in LLMs, there are millions and even billions of parameters. Our method is expected to be then more efficient in such cases.

\subsection{Convergence time}
As mentioned in the previous section, our ZO method requires a number of iterations $T>T'$ for gradient-based FL methods.  At first look, one would consider that the convergence time of standard FL is less than our method. This is actually the case if one neglects the upload time and the limited capacity of wireless links. For instance, in current wireless standards, the bit rate of each user is equal to a few Mbit/s. Let us consider an example where the bit rate is $10$Mbit/s. In each FL round/iteration, there are $64\times 10^5$ bits to upload in standard FL, which takes $640$ms. If we ignore the downlink transmission time and the computation time (for the gradient and the aggregation), and if $T'=100$, the total time is $64$sec. In fact, the convergence time bottleneck is due essentially to the upload time in gradient-based methods. In our method, this issue is solved, and the convergence time is no longer limited by the upload time of the wireless links since only one scalar is transmitted per iteration, which can be done in much less than $1$ms. For instance, one time slot is needed in our case, which can be equal to $0.125$ms in 5G. If $T=10000$, the total upload time is $1.25sec$. The convergence time of our method is, hence, essentially limited by the computation time of the loss function (forward propagation). One can see that for $d=4\times 10^5$ parameters, around a few million operations are required (summation of weighted inputs to each neuron and then performing the activation function and roughly there are less than $1000$ inputs per neuron and less than $1000$ neurons per layer since otherwise, the number of parameters would be much higher). Roughly speaking, if $16$ millions of operations are needed, and for a $4$GHz CPU, the computation time per iteration is $4$ms, and hence the total communication and computation time (convergence time) is around $41,25$sec, which is less than the communication time of standard FL method (without considering the computation time of standard FL).  This shows clearly that the convergence time of our method could be less than that of standard FL, even if the number of iterations required in our method is bigger.

\subsection{Energy Consumption}
The energy consumption in FL is the total sum of computation/processing energy and transmission energy. 
Since each device transmits a high amount of information, the transmission energy by the devices represents a central part of the total energy consumption in the system. In our method, only one quantized scalar of $16$bits is needed and the transmission energy required is hence negligible compared to the standard FL method. 
As for the processing energy for local training, it is divided between the energy used for inference (forward propagation (FP)) and that for the backward propagation (BP) (gradient computation). The processing energy is essentially due to the number of multiply-accumulate (MAC) operations, the precision level of the quantization, accessing the main memory (SRAM), and fetching data from the DRAM. The forward propagation energy  per iteration round is \cite{Ereference}, for $M$ quantization bits, 
\begin{align}
    & E_{FP}=E_{computing}+E_{w}+E_{b}+E_{DRAM} \\
    & E_{computing} = E_{MAC}(M) N_c+3 O_c E_{MAC}(M_{max}) \\ 
    & E_{w} = 2E_{MAC}(M) d +E_{MAC}(M) N_c\sqrt{\frac{M}{p_u M_{max}}} \\
    & E_{b}= 2E_{MAC}(M)O_c +E_{MAC}(M)N_c\sqrt{\frac{M}{p_u M_{max}}} \\
    & E_{DRAM} = A_d E_{MAC}(M_{max}) x_{in} \nonumber \\ &\hspace{2cm}+2 A_d E_{MAC}(M) \max\big(d M+O_c M-S,0\big)
\end{align}
where $E_w$ denotes the energy required to retrieve weights from the buffers, $E_b$ the energy to retrieve activations from the buffers, 
and $E_{DRAM}$ the energy to retrieve input features and weights from the DRAM. $M_{max}$ is the maximum precision level, $p$ is the number of MAC units, $N_c$ is the number of MAC operations, $O_c$ the total number of activations throughout the whole network, $x_{in}$  the input size, and S the SRAM buffer size. 
Following \cite{Ereference},
$E_{MAC}(M)=A\big(\frac{M}{M_{max}}\big)^{\mu}$, with $1<\mu<2$ and $A>0$. 

The backward propagation energy consumption per iteration/round is \cite{Kim}, 
\begin{align}
     E_{BP} & =2N_c E_{MAC}(M_{max}) +2O_c E_{MAC}(M_{max}) \nonumber \\ &+d E_{MAC}(M_{max}) + 2N_c E_{MAC}(M_{max})\sqrt{\frac{1}{p}} \nonumber \\ & + 2A_d E_{MAC}(M_{max}) max\big(d M_{max}+O_c M_{max}-s_m,0\big)
\end{align}
In standard FL, the total consumption energy is the summation of the total transmission energy, the total forward propagation energy ($E_{FP}\times T'$), and the total backward propagation energy ($E_{BP}\times T'$) for the total training time.  

In our DZOFL method, the backward propagation is not used since there is no need to compute the gradient. The total energy is, hence, the summation of the total transmission energy and the total forward propagation energy ($E_{FP}\times T$). It is worth mentioning that the transmission energy contains both the uplink and downlink transmission energies.

In Section \ref{num}, we will show that the total energy consumption in our method is lower than that of the standard FL.

\section{Convergence analysis}
This section analyzes the behavior of our algorithms in the nonconvex setting. 
Assuming that a global minimizer $\theta^*\in\mathbb{R}^{d}$ exists such that $\min_{\theta\in\mathbb{R}^d} F(\theta) = F(\theta^*)>-\infty$ and $\nabla F(\theta^*)=0$, we start by introducing necessary assumption on the global objective function.

\begin{assumption}\label{objective_fct} 
	We assume the existence and the continuity of $\nabla F_i(\theta)$ and $\nabla^2 F_i(\theta)$, and that there exists a constant $\alpha_1>0$ such that
	$\|\nabla^2 F_i(\theta)\|_2\leq\alpha_1$,$\forall i\in\mathcal{N}$.	
\end{assumption}
\begin{assumption}\label{local_fcts}
	All loss functions $\theta\mapsto f_i(\theta,\xi_i)$ are Lipschitz continuous with Lipschitz constant $L_{\xi_i}$,
	$|f_i(\theta,\xi_i)-f_i(\theta',\xi_i)|\leq L_{\xi_i}\|\theta-\theta'\|$, $\forall i\in\mathcal{N}$.
	In addition, $\mathbb{E}_{\xi_i} f_i (\theta, \xi_i) < \infty, \forall i \in\mathcal{N}$.
\end{assumption}
We also consider standard assumptions about the step sizes. 
\begin{assumption}\label{step_sizes_1}
	Both the step sizes $\alpha_k$ and $\gamma_k$ vanish to zero as $k\rightarrow\infty$ and the following series composed of them satisfy the convergence assumptions
	$\sum_{k=0}^{\infty}\alpha_k \gamma_k = \infty$, $\sum_{k=0}^{\infty}\alpha_k\gamma_k^3 <\infty$, and $\sum_{k=0}^{\infty}\alpha_k^2\gamma_k^2<\infty$.
\end{assumption}
\begin{example}\label{eg}
	To satisfy Assumption \ref{step_sizes_1}, we consider the following form of the step sizes, $\alpha_k = \alpha_0(1+k)^{-\upsilon_1}$ and $\gamma_k = \gamma_0 (1+k)^{-\upsilon_2}$ with $\upsilon_1, \upsilon_2>0$. Then, it's sufficient to find $\upsilon_1$ and $\upsilon_2$ such that $0<\upsilon_1+\upsilon_2\leq 1$, $\upsilon_1+3\upsilon_2>1$, and $\upsilon_1+\upsilon_2>0.5$.
\end{example}
In addition, we consider the following standard assumption about the quantizer. 
\begin{assumption}\label{quantizer}
	The random quantizer $Q(\cdot)$ is unbiased and its variance grows with the square of l2-norm of its argument, i.e.,
	\begin{equation}
		\mathbb{E}[Q(x)|x]=x \;\;\;\text{and}\;\;\; \mathbb{E}[\|Q(x)-x\|^2|x]\leq \sigma \|x\|^2,
	\end{equation}
	for some real positive constant $\sigma$ and any $x\in\mathbb{R}^d$.
\end{assumption}

\begin{lemma}\label{smooth}
	By Assumption \ref{objective_fct}, we know that the objective function $\theta\longmapsto F(\theta)$ is $L$-smooth for some positive constant $L$,
	$\|\nabla F(\theta)-\nabla F(\theta')\|\leq L\|\theta-\theta'\|, \;\forall \theta,\theta'\in\mathbb{R}^d,$	or equivalently, 
	$F(\theta)\leq F(\theta')+\langle\nabla F(\theta'), \theta-\theta'\rangle +\frac{L}{2}\|\theta-\theta'\|^2$.
	
\end{lemma}

\begin{assumption}\label{perturbation}
	Let $\Phi_{k} = (\phi_{k}^1, \phi_{k}^2, \ldots, \phi_{k}^d)^T$.
	At each iteration $k$, the generated $\Phi_{k}$ vector is independent of other iterations. In addition, the elements of $\Phi_{k}$ are assumed i.i.d with $\mathbb{E}(\phi_{k}^{d_1} \phi_{k}^{d_2}) =0$ for $d_1 \neq d_2$ and there exists $\alpha_2 >0$ such that
	$\mathbb{E} (\phi_{k}^{d_j})^2 = \alpha_2$, $\forall {d_j}$, $\forall k$.
	We further assume there exists a constant $\alpha_3 >0$ where
	$\|\Phi_{k}\|\leq \alpha_3$, $\forall k$.
\end{assumption}
\begin{example}\label{phi_eg}
	An example of a perturbation vector satisfying Assumption \ref{perturbation}, is picking every dimension of $\Phi_{k}$ from $\{-\frac{1}{\sqrt{d}},\frac{1}{\sqrt{d}}\}$ with equal probability. Then, $\alpha_2=\frac{1}{d}$ and $\alpha_3=1$.
\end{example}

We now provide the convergence of our algorithm in the following theorem.

The proof of convergence follows in several steps. First, we prove that our method allows obtaining, on average, a biased estimation of the gradient.



	

The smoothness inequality allows for the first main result, leading to the second in the following theorem.
\begin{theorem}\label{th-ncvx}
	When Assumptions \ref{objective_fct}-\ref{perturbation} hold, we have $\sum_{k}\alpha_k\gamma_k\mathbb{E}[\|\nabla F(\theta_k)\|^2]< +\infty$ and $\lim_{k\rightarrow\infty}\mathbb{E}[\|\nabla F(\theta_k)\|^2]=0$, meaning that the algorithm converges.
	
	Proof: Refer to Appendices A and \ref{th-ncvx-proof}.
\end{theorem}
Proof sketch: The proof follows in several steps. First, we analyze the term $\mathbb{E}(g_k)$ in Appendix A and show it is a biased (not unbiased) estimate of the gradient. This adds difficulties in proving the convergence since our method deviates from the standard gradient method. We then substitute the algorithm's updates in the second inequality of Lemma \ref{smooth} and study the conditional expectation given the history sequence. We then perform a recursive addition over the iterations $k>0$. With the conditions on the step sizes and the upper bound on the estimate's and its bias' squared norm, we are able to find an upper bound on $\sum_k\alpha_k\gamma_k\mathbb{E}\big(\|\nabla F(\theta_k)\|^2\big)$ when $k$ grows to $\infty$. The next step is to consider the hypothesis $\lim_{k\rightarrow\infty}\mathbb{E}\big(\|\nabla F(\theta_k)\|\big)\geq \rho$, for $\rho>0$, and prove that it contradicts with the first result.

 We then provide an upper bound on the convergence rate of Algorithm \ref{alg:example_2p}.
\begin{theorem}\label{th-ncvx-rate}
	In addition to the assumptions of Theorem \ref{th-ncvx}, let the step sizes have the form of Example \ref{eg} with $\upsilon_3=\upsilon_1+\upsilon_2<1$. Then, 
	
	\begin{equation}\label{f_1}
		\begin{split}
			&\frac{\sum_{k}\alpha_k\gamma_k\mathbb{E}\big[\|\nabla F(\theta_k)\|^2\big]}{\sum_k \alpha_k\gamma_k}	
			\leq
			\\&\frac{(1-\upsilon_3)}{(K+2)^{1-\upsilon_3}-1}\bigg(A_0+ \frac{A_1}{\upsilon_1+3\upsilon_2-1}+\frac{A_2}{2\upsilon_3-1}\bigg).\\
		\end{split}
	\end{equation}
	
	where $A_0=\frac{2\delta_0}{c_1\alpha_0\gamma_0}$, $A_1=(\upsilon_1+3\upsilon_2)(c_3\gamma_0)^2$, and $A_2=\frac{2\upsilon_3c_2\alpha_0\gamma_0L}{c_1}$. 
	
	Proof: Refer to Appendix \ref{th-ncvx-rate-proof}.
\end{theorem}
In Theorem \ref{th-ncvx-rate}, the optimal choice of exponents in equation (\ref{f_1}) is $\upsilon_1=\upsilon_2=\frac14$, resulting in a rate of $O\left(\frac{1}{\sqrt{K}}\right)$. However, to prevent the constant part from becoming excessively large, we identify a very small value $\epsilon > 0$ such that $\upsilon_1=\upsilon_2=\frac14+\frac{\epsilon}{2}$, leading to a rate of $O\left(\frac{1}{K^{\frac12-\epsilon}}\right)$. This result is remarkable as the obtained rate competes with standard gradient methods that require the exchange of long vectors of gradients, while in our method, only a scalar is required to be sent by each device, resulting hence in a huge saving of communication overhead and energy consumption.

\section{Numerical Results}\label{num}
We consider an FL setting where 50 devices, randomly placed in a cell of radius 500m. The bandwidth allocated to each device is 2MHz. We used a standard channel model with Rayleigh fast fading. The gaussian noise is -173dBm/Hz.

The goal of the training is binary image classification from the FashionMNIST data set where we consider that the data is distributed among the devices in an i.i.d. manner. We consider a CNN with two convolutional layers composed of $20$ kernels and $40$ kernels, both of size $7\times 7$. The second layer is followed by $2\times 2$ max pooling, then a flattening layer, and a final linear layer of $2$ outputs. We use ReLU activations and we use batches of size $10$. For this example, we have $N_c = 10.56\times 10^6$, $O_c= 25042\times 10$, $d=45362$, $S=8MB$, $x_{in}=784$, $M=16$, $M_{max}=32$, A=$3.7$pJ, $A_d$=150, $\mu=1.25$, $p=64$.  
The data is randomly and equally distributed among the users. We classify images with labels "shirt" and "sneaker" and test the accuracy against an independent test set at each communication round. 

We show in this section the total energy consumption (for transmission and computation) as well as the total convergence time. 
The results show that our algorithm achieves better convergence time and energy consumption. In fact, even if the number of iterations in DZOFL is bigger than standard FL, the high amount of information to be transmitted over the wireless links requires a non negligible duration which makes the duration of each round/iteration big. In our case, only one symbol needs to be transmitted by each user which can be done in one slot. In fact, in standard FL, the main training time is essentially due to the transmission time over the wireless links, while the computation time is non negligible but can be much less than the transmission time. In DZOFL, the main delay is essentially due to the computation time, which turns out to be lower than the convergence time of standard FL, although higher number of iterations is required for convergence. 

Regarding the energy consumption, one can explain the results using the same reasons. In standard FL, there is a high energy needed for the transmission, while in our case the main energy consumption is due to the computation and the transmission energy is negligible.  
\begin{figure}
    \centering
    \includegraphics[width=0.95\linewidth]{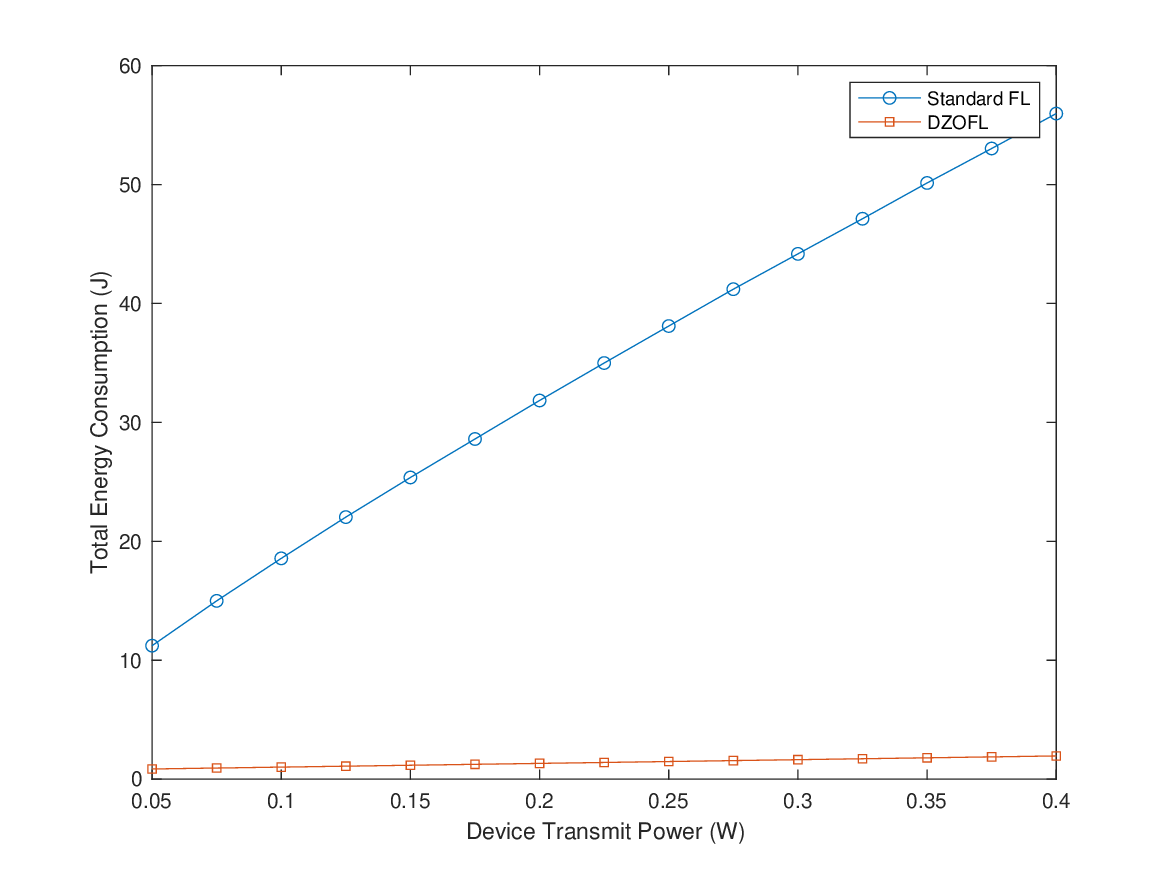}
    \caption{Total Energy Consumption (joule)}
    \label{fig:energy-32}
\end{figure}
\begin{figure}
    \centering
    \includegraphics[width=0.95\linewidth]{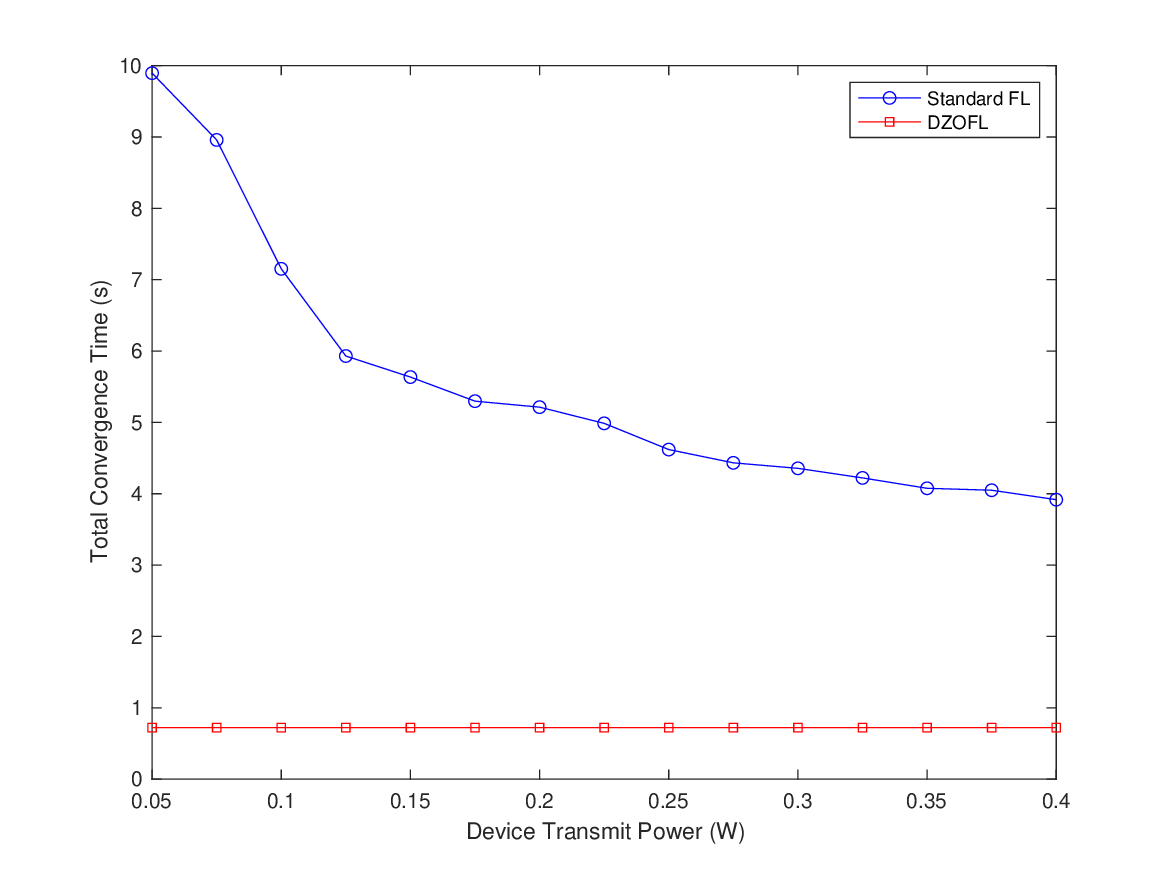}
    \caption{Total Convergence Time (sec)}
    \label{fig:convergencetime-32}
\end{figure}

\section{Conclusion}
This paper studied the communication and energy consumption bottlenecks in FL. We considered a wireless setting where the devices collectively train a model using FL over wireless links, and hence, the information sent over the links is subject to errors and packet dropping. We developed a ZO-based FL method where each device sends only one scalar instead of a long gradient vector as in standard FL. Furthermore, the impact of quantization and packet dropping due to wireless errors is also considered in our method. We proved the convergence of our method in a non-convex setting and provided its convergence rate that competes with standard gradient techniques while requiring much less communication overhead. We have then shown the superiority of our method in terms of energy consumption and communication overhead.   

\appendices

\section{Analysis of $\mathbb{E}(g_k)$}
In this section we prove useful results that will be used in the convergence proof. 
As we deal with stochastic environments, we inevitably analyze the expectation over all possible variable outcomes. From Lemma \ref{biased_estimators}, we see that in expectation, our estimator deviates from the gradient direction by the bias term. We then prove that  this term does not grow larger and even grows smaller as the algorithms evolves. Additionally, to ensure that the expected norm squared of the estimator, as shown in Lemma \ref{norm}, does not accumulate residual constant terms. 

The results here shows clearly that our method is different from the stochastic gradient technique. In fact, a stochastic gradient is an unbiased estimation of the gradient, while here the expectation of $\mathbb{E}(g_k)$ is a biased estimation of the gradient. The bias term is not equal to zero and it implies difficulties hence in proving the convergence. 

\begin{equation}\label{2p_grdt_est2}
	\begin{split}
		g_k = & \Phi_k \times Q(\Delta f_k)\\
		= &\Phi_k Q\Bigg(\frac{N}{|\mathcal{S}_k|} \sum_{i\in\mathcal{S}_k} Q\Big(f_i\big(\theta_k + \gamma_k\Phi_k, \xi_{i,k}\big)\\&\hspace{3cm}-f_i\Big(\theta_k - \gamma_k\Phi_k, \xi_{i,k}\big)\Big)\Bigg)
	\end{split}
\end{equation}

Let $\mathcal{H}_k = \{\theta_0, \xi_0, \theta_1, \xi_1, ..., \theta_{k-1}, \xi_{k-1}, \theta_k\}$ denote the history sequence, then the following two Lemmas characterize our gradient estimates. 
\begin{lemma}\label{lemma-erasure}
	Let Assumption  \ref{quantizer} holds. The expectation of $\Delta f_k$ over all possible sets $\mathcal{S}_k$ and the random quantization $Q(\cdot)$ is proportional to $\sum_{i\in\mathcal{N}}\Delta f_{i,k}$, i.e,
	\begin{equation}
		\mathbb{E}_{\mathcal{S}_k, Q}\big(\Delta f_k\big|\mathcal{H}_k, \Phi_k, \xi_k\big) =  q \sum_{i\in\mathcal{N}} \Delta f_{i,k}, \;\;\; \forall k>0,
	\end{equation}
	with the constant $q=\mathbb{P}(|\mathcal{S}_k|\neq 0) = 1-(1-p)^N$.
	
	Proof: Refer to Appendix \ref{e-q}.
\end{lemma}

\begin{lemma}\label{biased_estimators}
	Let Assumptions \ref{objective_fct}-\ref{perturbation} be satisfied and define the scalar value $c_1=2q \alpha_2$. Then, the gradient estimator is biased w.r.t. the objective function's exact gradient $\nabla F(\theta)$. Concretely, 		
	$$\mathbb{E}[g_k|\mathcal{H}_k] = c_1\gamma_k(\nabla F(\theta_k)+b_k),$$ 
	where $b_k$ is the bias term.
	
	Proof: Refer to Appendix \ref{app-grdt_est}.
	
	\begin{lemma}\label{norm}
		Let Assumptions \ref{objective_fct}-\ref{perturbation} hold and define the scalar value $c_2=4q(\sigma+1)^2\alpha_3^4 N^2L_{\xi}$, where $L_{\xi} = \max_i \mathbb{E}[L_{\xi_{i,k}}^2|\mathcal{H}_k]$. Then,
		$$\mathbb{E}[\|g_k\|^2|\mathcal{H}_k] \leq c_2\gamma_k^2.$$ 
		
		Proof: Refer to Appendix \ref{norm-sq}.
	\end{lemma}
\end{lemma}

\begin{lemma}\label{bias-norm-lemma}
	By Assumptions \ref{perturbation} and \ref{objective_fct}, we can find a scalar value $c_3>0$ such that $$\|b_k\| \leq c_3\gamma_k.$$
	
	Proof: Refer to Appendix \ref{bias-norm}.
\end{lemma}

\subsection{Proof of Lemma \ref{lemma-erasure}: Effect of the Packet Erasure Channel}\label{e-q}
The conditional expectation based on (\ref{reception}) given $\mathcal{H}_k$ and the cardinal of $\mathcal{S}_k$ is written as 
\begin{equation}\label{bigpart}
	\begin{split}
		&\mathbb{E}_{\mathcal{S}_k, Q}\big(\Delta f_k\big| |\mathcal{S}_k|=n, \mathcal{H}_k, \Phi_k, \xi_k\big) \\ &= 
		\begin{cases}
			\frac{N}{n}\mathbb{E}_{\mathcal{S}_k}\Big(\sum_{i\in\mathcal{S}_k} \Delta f_{i,k}\Big| |\mathcal{S}_k|=n, \mathcal{H}_k, \Phi_k, \xi_k\Big), &\text{if } n\neq 0,\\
			0, &\text{if } n= 0,
		\end{cases}
	\end{split}
\end{equation}
where we used the fact that the quantization is unbiased.

Let $\mathcal{I}^{(n)}$ be the collection of all possible sets $\mathcal{S}_k$ such that $|\mathcal{S}_k|=n$. For example, $\mathcal{I}^{(1)}=\{\{1\},\{2\},\ldots,\{N\}\}$. Knowing that each user has an equal probability of participating in $\mathcal{S}_k$ and its participation is independent of others, the probability of selecting any specific combination of $n$ users is the same, i.e., the sets in $\mathcal{I}^{(n)}$ are equiprobable.
We also know that $|\mathcal{I}^{(n)}|=\binom{N}{n}$. Thus,
\begin{equation}
	\mathbb{P}\Big(\mathcal{S}_k=\mathcal{M}\Big| |\mathcal{S}_k|=n\Big) = \frac{1}{\binom{N}{n}}, \;\;\;\forall\mathcal{M}\in\mathcal{I}^{(n)},
\end{equation}
noting that $\mathcal{S}_k$ is independent of $\mathcal{H}_k$, $\Phi_k$, and $\xi_k$. 

Hence,
\begin{equation}\label{subpart}
	\begin{split}
		&\mathbb{E}_{\mathcal{S}_k}\Big(\sum_{i\in\mathcal{S}_k} \Delta f_{i,k}\Big| |\mathcal{S}_k|=n, \mathcal{H}_k, \Phi_k, \xi_k\Big)\\ = &\sum_{\mathcal{M}\in\mathcal{I}^{(n)}}\frac{1}{\binom{N}{n}}\sum_{i\in\mathcal{M}}\Delta f_{i,k} \\
		= &\frac{1}{\binom{N}{n}}\sum_{\mathcal{M}\in\mathcal{I}^{(n)}}\sum_{i\in\mathcal{M}}\Delta f_{i,k} \\
		\overset{(a)}{=} &\frac{1}{\binom{N}{n}}\binom{N-1}{n-1}\sum_{i\in\mathcal{N}}\Delta f_{i,k} \\
		=&\frac{n}{N}\sum_{i\in\mathcal{N}}\Delta f_{i,k},
	\end{split}
\end{equation}
where in $(a)$, we know that, $\forall i\in\mathcal{N}$, $\Delta f_{i,k}$ appears in the previous double sum as many times as user $i$ appears in the sets of $\mathcal{I}^{(n)}$. To find out how many of the combinations in $\mathcal{I}^{(n)}$ include user $i$, we need to consider how many ways we can choose the remaining $n-1$ users from the remaining $N-1$ users (excluding user $i$). The number of ways is equal to $\binom{N-1}{n-1}$.

Combining (\ref{bigpart}) and (\ref{subpart}) for any $1\leq n\leq N$, we obtain
\begin{equation}
	\mathbb{E}_{\mathcal{S}_k, Q}\big(\Delta f_k\big| |\mathcal{S}_k|=n, \mathcal{H}_k, \Phi_k, \xi_k\big) = \sum_{i\in\mathcal{N}}\Delta f_{i,k}.
\end{equation}

Computing the full expectation,
\begin{equation}\label{fullexp}
	\begin{split}
		&\mathbb{E}_{\mathcal{S}_k, Q}\big(\Delta f_k\big|\mathcal{H}_k, \Phi_k, \xi_k\big) \\ = &\sum_{n=0}^{N}\mathbb{P}(|\mathcal{S}_k|=n)\mathbb{E}_{\mathcal{S}_k, Q}\big(\Delta f_k\big| |\mathcal{S}_k|=n, \mathcal{H}_k, \Phi_k, \xi_k\big)\\
		= &\sum_{n=1}^{N}\mathbb{P}(|\mathcal{S}_k|=n)\sum_{i\in\mathcal{N}}\Delta f_{i,k}\\
		= &\big(1-\mathbb{P}(|\mathcal{S}_k|=0)\big)\sum_{i\in\mathcal{N}}\Delta f_{i,k}\\
		= &q\sum_{i\in\mathcal{N}}\Delta f_{i,k}.
	\end{split}
\end{equation}

\subsection{Proof of Lemma \ref{biased_estimators}: Biased Estimator}\label{app-grdt_est}
let $g_k$ have the form in (\ref{2p_grdt_est}), then
 
\begin{equation*}
	\begin{split}
		&\mathbb{E}[g_k|\mathcal{H}_k] \\
		\overset{(a)}{=} &\mathbb{E}\Big[\mathbb{E}\big[\Phi_k Q(\Delta f_k) \big|\mathcal{H}_k,\Phi_k, \Delta f_k \big]\Big|\mathcal{H}_k\Big]\\
        \overset{(b)}{=} &\mathbb{E}\Big[\Phi_k \Delta f_k\Big|\mathcal{H}_k\Big]\\
        = &\mathbb{E}_{\Phi_k,\xi_k,\mathcal{S}_k,Q}\Big[\Phi_k \Delta f_k\Big|\mathcal{H}_k\Big]\\
		= &\mathbb{E}_{\Phi_k,\xi_k}\bigg[\mathbb{E}_{\mathcal{S}_k,Q}\Big[\Phi_k \Delta f_k \Big|\mathcal{H}_k,\Phi_k,\xi_k\Big]\Big|\mathcal{H}_k\bigg]\\
		\overset{(c)}{=} &q\mathbb{E}_{\Phi_k,\xi_k}\bigg[\Phi_k\sum_{i\in\mathcal{N}} \Delta f_{i,k} \Big|\mathcal{H}_k\bigg]\\
		= &q\mathbb{E}_{\Phi_k,\xi_k}\bigg[\Phi_k\sum_{i=1}^{N}\Big[f_i\big(\theta_k + \gamma_k\Phi_k, \xi_{i,k}\big)\\&\hspace{3cm}-f_i\Big(\theta_k - \gamma_k\Phi_k, \xi_{i,k}\big)\Big] \Big|\mathcal{H}_k\bigg]\\
		\overset{(d)}{=} &q\mathbb{E}_{\Phi_k}\bigg[\Phi_k \sum_{i=1}^{N}\Big[F_i\Big(\theta_k + \gamma_k\Phi_k\Big)-F_i\Big(\theta_k - \gamma_k\Phi_k\Big)\Big]\Big|\mathcal{H}_k\bigg]\\
		\overset{(e)}{=} &q\mathbb{E}_{\Phi_k}\bigg[\Phi_k \sum_{i=1}^{N}\Big[F_i(\theta_k)+\gamma_k\Phi_k^T\nabla F_i(\theta_k)
		+\frac{\gamma_k^2}{2}\Phi_k^T \nabla^2 F_i(\acute{\theta}_k)\Phi_k\\&-\big(F_i(\theta_k)-\gamma_k\Phi_k^T\nabla F_i(\theta_k)
		+\frac{\gamma_k^2}{2} \Phi_k^T \nabla^2 F_i(\grave{\theta}_k)\Phi_k\big) \Big]\Big|\mathcal{H}_k\bigg]\\
        = &q\mathbb{E}_{\Phi_k}\bigg[\Phi_k    \sum_{i=1}^{N}\bigg(2\gamma_k\Phi_k^T\nabla F_i(\theta_k)\\&\hspace{2cm}+\frac{\gamma_k^2}{2}\Phi_k^T (\nabla^2 F_i(\acute{\theta}_k)-\nabla^2F_i(\grave{\theta}_k))\Phi_k \bigg)\Big|\mathcal{H}_k\bigg]\\
        = &2q\gamma_k\mathbb{E}_{\Phi_k}\bigg[ \sum_{i=1}^{N}\Phi_k\Phi_k^T\nabla F_i(\theta_k)|\mathcal{H}_k\bigg]\\&+q\frac{\gamma_k^2}{2} \mathbb{E}_{\Phi_k}\bigg[ \sum_{i=1}^{N} \Phi_k\Phi_k^T (\nabla^2 F_i(\acute{\theta}_k)-\nabla^2F_i(\grave{\theta}_k))\Phi_k\Big|\mathcal{H}_k\bigg]\\
        = &2q\gamma_k \sum_{i=1}^{N}\mathbb{E}_{\Phi_k}\big[\Phi_k \Phi_k^T\big|\mathcal{H}_k\big]\nabla F_i(\theta_k)\\&+q\frac{\gamma_k^2}{2}\sum_{i=1}^{N}\mathbb{E}_{\Phi_k}\bigg[\Phi_k\Phi_k^T (\nabla^2 F_i(\acute{\theta}_k)-\nabla^2F_i(\grave{\theta}_k))\Phi_k\Big|\mathcal{H}_k\bigg]\\
      \end{split}
    \end{equation*}
    \begin{equation}\label{2p_grdt_exp}
    	\begin{split}
		\overset{(f)}{=} &2q\alpha_2\gamma_k\sum_{i=1}^{N} \nabla F_i(\theta_k)\\&+q\frac{\gamma_k^2}{2}\sum_{i=1}^{N}\mathbb{E}_{\Phi_k}\bigg[\Phi_k\Phi_k^T (\nabla^2 F_i(\acute{\theta}_k)-\nabla^2F_i(\grave{\theta}_k))\Phi_k\Big|\mathcal{H}_k\bigg]\\
		\overset{(g)}{=} &c_1 \gamma_k(\nabla F(\theta_k)+b_k)\\
	\end{split}
\end{equation}
where $(a)$ is due to the law of total expectation, $(b)$ is by the unbiasedness of the quantizer, $(c)$ is due to Lemma \ref{lemma-erasure}, $(d)$ is by the definition in (\ref{F_i}), $(e)$ is by Taylor expansion and mean-valued theorem and considering $\acute{\theta}_k$ between $\theta_k$ and $\theta_k + \gamma_k\Phi_k$, and $\grave{\theta}_k$ between $\theta_k$ and $\theta_k -\gamma_k\Phi_k$. $(f)$ is due to Assumption \ref{perturbation}. In $(g)$,  we let $c_1=2q \alpha_2 $.

From (\ref{2p_grdt_exp}), we can see that the estimate bias has the form  
\begin{equation}\label{2p_bias}
	b_k = \frac{\gamma_k}{4\alpha_2}\sum_{i=1}^{N}\mathbb{E}\bigg[\Phi_k\Phi_k^T (\nabla^2 F_i(\acute{\theta}_k)-\nabla^2F_i(\grave{\theta}_k))\Phi_k\Big|\mathcal{H}_k\bigg].
\end{equation}
\subsection{Proof of Lemma \ref{norm}: Expected Norm Squared of the Estimated Gradient}\label{norm-sq}
Bounding the norm squared of the gradient estimate,
\begin{equation*}
	\begin{split}
		&\mathbb{E}_{\Phi_k,\xi_k,\mathcal{S}_k,Q}[\|g_k\|^2|\mathcal{H}_k]\\ 
		= &\mathbb{E}_{\Phi_k,\xi_k,\mathcal{S}_k,Q}\bigg[\|\Phi_k Q(\Delta f_k)\|^2\Big|\mathcal{H}_k\bigg]\\
		\overset{(a)}{\leq}  &\alpha_3^2\mathbb{E}_{\Phi_k,\xi_k,\mathcal{S}_k,Q}\bigg[ Q^2(\Delta f_k)\Big|\mathcal{H}_k\bigg]\\
            = &\alpha_3^2\mathbb{E}_{\Phi_k,\xi_k,\mathcal{S}_k,Q}\bigg[ \Big(Q(\Delta f_k)-\Delta f_k+\Delta f_k\Big)^2\Big|\mathcal{H}_k\bigg]\\
           = &\alpha_3^2 \mathbb{E}_{\Phi_k,\xi_k,\mathcal{S}_k}\Bigg[ \mathbb{E}_{Q}\Bigg[ \Big[\Big(Q(\Delta f_k)-\Delta f_k\Big)^2+\Big(\Delta f_k\Big)^2\\&\hspace{0.5cm}+2\Big(Q(\Delta f_k)-\Delta f_k\Big)\cdot\Delta f_k \Big]\Big|\mathcal{H}_k,\Phi_k,\xi_k,\mathcal{S}_k\Bigg]\Big|\mathcal{H}_k\Bigg]\\
           \overset{(b)}{\leq} &\alpha_3^2 \mathbb{E}_{\Phi_k,\xi_k,\mathcal{S}_k, Q}\Bigg[\sigma \Big(\Delta f_k\Big)^2 +\Big(\Delta f_k\Big)^2 \Big|\mathcal{H}_k\Bigg]\\
            = &\alpha_3^2 (\sigma+1) \mathbb{E}_{\Phi_k,\xi_k,\mathcal{S}_k, Q}\Big[\Big(\Delta f_k\Big)^2\Big|\mathcal{H}_k\Big]\\
		= &\alpha_3^2(\sigma+1)\mathbb{E}_{\Phi_k,\xi_k,\mathcal{S}_k,Q}\bigg[\Big( \frac{N}{|\mathcal{S}_k|}\underset{i\in\mathcal{S}_k}{\sum} Q(\Delta f_{i,k})\Big)^2\Big|\mathcal{H}_k\bigg]\\
		\leq &\alpha_3^2 (\sigma+1)\mathbb{E}_{\Phi_k,\xi_k,\mathcal{S}_k,Q}\bigg[  \frac{N^2}{|\mathcal{S}_k|^2}\Big(\underset{i\in\mathcal{S}_k}{\sum} Q(\Delta f_{i,k})\Big)^2\Big|\mathcal{H}_k\bigg]\\
        \overset{(c)}{\leq} &\alpha_3^2 (\sigma+1)N^2\mathbb{E}_{\Phi_k,\xi_k,\mathcal{S}_k,Q}\bigg[  \frac{1}{|\mathcal{S}_k|}\underset{i\in\mathcal{S}_k}{\sum} Q^2(\Delta f_{i,k})\Big|\mathcal{H}_k\bigg]\\
		= &\alpha_3^2 (\sigma+1) N^2\times\\
		&\mathbb{E}_{\Phi_k,\xi_k,\mathcal{S}_k,Q}\bigg[  \frac{1}{|\mathcal{S}_k|}\underset{i\in\mathcal{S}_k}{\sum} \Big(Q(\Delta f_{i,k})-\Delta f_{i,k}+\Delta f_{i,k}\Big)^2\Big|\mathcal{H}_k\bigg]\\
		= &\alpha_3^2 (\sigma+1)N^2\mathbb{E}_{\Phi_k,\xi_k,\mathcal{S}_k,Q}\bigg[  \frac{1}{|\mathcal{S}_k|}\underset{i\in\mathcal{S}_k}{\sum} \Big[\Big(Q(\Delta f_{i,k})-\Delta f_{i,k}\Big)^2\\&\hspace{1cm}+\Big(\Delta f_{i,k}\Big)^2+2\Big(Q(\Delta f_{i,k})-\Delta f_{i,k}\Big)\cdot\Delta f_{i,k}\Big]\Big|\mathcal{H}_k\bigg]\\
  \end{split}
\end{equation*}
  \begin{equation}\label{grdt_norm}
	\begin{split}
        = &\alpha_3^2 (\sigma+1) N^2\times\\
		&\mathbb{E}_{\Phi_k,\xi_k,\mathcal{S}_k}\Bigg[ \mathbb{E}_{Q}\Bigg[  \frac{1}{|\mathcal{S}_k|}\underset{i\in\mathcal{S}_k}{\sum}  \Big[\Big(Q(\Delta f_{i,k})-\Delta f_{i,k}\Big)^2+\Big(\Delta f_{i,k}\Big)^2\\&\hspace{0.5cm}+2\Big(Q(\Delta f_{i,k})-\Delta f_{i,k}\Big)\cdot\Delta f_{i,k} \Big]\Big|\mathcal{H}_k,\Phi_k,\xi_k,\mathcal{S}_k\Bigg]\Big|\mathcal{H}_k\Bigg]\\
        \leq &\alpha_3^2 (\sigma+1)N^2\times \\ &\mathbb{E}_{\Phi_k,\xi_k,\mathcal{S}_k}\Bigg[  \frac{1}{|\mathcal{S}_k|}\underset{i\in\mathcal{S}_k}{\sum}\sigma\Big(\Delta f_{i,k}\Big)^2+\Big(\Delta f_{i,k}\Big)^2\Big|\mathcal{H}_k\Bigg]\\
		= &\alpha_3^2 N^2(\sigma+1)^2\mathbb{E}_{\Phi_k,\xi_k,\mathcal{S}_k}\Bigg[  \frac{1}{|\mathcal{S}_k|}\underset{i\in\mathcal{S}_k}{\sum}\big(\Delta f_{i,k}\big)^2\Big|\mathcal{H}_k\Bigg]\\
        \overset{(d)}{=} &\alpha_3^2 N^2(\sigma+1)^2\mathbb{E}_{\Phi_k,\xi_k}\Bigg[  \frac{q}{N}\underset{i\in\mathcal{N}}{\sum}\big(\Delta f_{i,k}\big)^2\Big|\mathcal{H}_k\Bigg]\\
        = &q\alpha_3^2 N(\sigma+1)^2\mathbb{E}_{\Phi_k,\xi_k}\Bigg[  \underset{i\in\mathcal{N}}{\sum}\Big(f_i\Big(\theta_k + \gamma_k\Phi_k, \xi_{i,k}\Big)\\&\hspace{4cm}-f_i\Big(\theta_k - \gamma_k\Phi_k, \xi_{i,k}\Big)\Big)^2\Big|\mathcal{H}_k\Bigg]\\
		\overset{(e)}{\leq} &q\alpha_3^2 N(\sigma+1)^2\mathbb{E}_{\Phi_k,\xi_k}\Bigg[  \underset{i\in\mathcal{N}}{\sum}L_{\xi_{i,k}}^2\|2 \gamma_k\Phi_k\|^2\Big|\mathcal{H}_k\Bigg]\\
		\overset{(f)}{\leq} &4q(\sigma+1)^2\alpha_3^4 N^2L_{\xi} \gamma_k^2  \\
		\overset{(g)}{=} &c_2 \gamma_k^2
  	\end{split}
\end{equation}
where $(a)$ is by Assumption \ref{perturbation} and $(b)$ is by Assumption \ref{quantizer}, where the first term is the variance of the quantization and is bounded above by $\sigma$ and the third term is zero due to the unbiasedness of the quantization.  $(c)$ is by Cauchy-Schwartz, $(\sum_{i=1}^{S}x_i)^2=(\sum_{i=1}^{S}1\cdot x_i)^2\leq S \sum_{i=1}^{S}x_i^2$. $(d)$ is by following similar steps leading up to (\ref{fullexp}). $(e)$ is by Assumption \ref{local_fcts}. In $(f)$, $L_{\xi} = \max_i \mathbb{E}[L_{\xi_{i,k}}^2|\mathcal{H}_k]$, and in $(g)$,  $c_2=4q(\sigma+1)^2\alpha_3^4 N^2L_{\xi}$.
\subsection{Proof of Lemma \ref{bias-norm-lemma}: Norm of the bias} \label{bias-norm}
The bias of (\ref{2p_bias}) can be bounded from above using Assumptions \ref{perturbation} and \ref{objective_fct}, as
\begin{equation}
	\begin{split}
		&\|b_k\| \\
		\overset{(a)}{\leq} &\frac{\gamma_k}{4\alpha_2}\sum_{i=1}^{N}\mathbb{E}\bigg[\|\Phi_k\| \|\Phi_k^T\| \|\nabla^2 F_i(\acute{\theta}_k)-\nabla^2 F_i(\grave{\theta}_k)\|\|\Phi_k\|\Big|\mathcal{H}_k\bigg]\\
		\overset{(b)}{\leq} &\frac{\alpha_1\alpha_3^3 N}{2\alpha_2}\gamma_k\\
		\overset{(c)}{=} &c_3\gamma_k,\\
	\end{split}
\end{equation}

where $(a)$ is due to Jensen's inequality, $(b)$ is due to Assumptions \ref{perturbation} and \ref{objective_fct}, and in $(c)$, $c_3=\frac{\alpha_1\alpha_3^3 N}{2\alpha_2}$.

\section{DZOFL Algorithm Convergence}

\subsection{Proof of Theorem \ref{th-ncvx}: Convergence analysis}\label{th-ncvx-proof}
Considering  the $L$-smoothness inequality applied to function $F$, we have
\begin{equation}
	\begin{split}
		F(\theta_{k+1})
		&\leq F(\theta_k)+\langle\nabla F(\theta_k), \theta_{k+1}-\theta_k\rangle +\frac{ L}{2}\|\theta_{k+1}-\theta_k\|^2.\\
	\end{split}
\end{equation}
which implies,
\begin{equation}
	\begin{split}
		F(\theta_{k+1})
		&\leq F(\theta_k)-\alpha_k\langle\nabla F(\theta_k), g_k\rangle +\frac{\alpha_k^2 L}{2}\|g_k\|^2.\\
	\end{split}
\end{equation}
Taking the conditional expectation given $\mathcal{H}_k$,
\begin{equation}\label{ncvx-rate1}
	\begin{split}
		&F(\theta_{k+1}) \\
		\overset{(a)}{\leq} &F(\theta_k)-c_1\alpha_k\gamma_k\langle\nabla F(\theta_k), \nabla F(\theta_k)+b_k\rangle +\frac{c_2L}{2}\alpha_k^2 \gamma_k^2\\
		= &F(\theta_k)-c_1 \alpha_k\gamma_k\|\nabla F(\theta_k)\|^2-c_1 \alpha_k\gamma_k\langle\nabla F(\theta_k), b_k\rangle\\&+\frac{c_2L}{2}\alpha_k^2 \gamma_k^2\\
		\overset{(b)}{\leq} &F(\theta_k)-c_1 \alpha_k\gamma_k\|\nabla F(\theta_k)\|^2+\frac{c_1 \alpha_k\gamma_k}{2}\|\nabla F(\theta_k)\|^2 \\&+\frac{c_1 \alpha_k\gamma_k}{2}\|b_k\|^2+\frac{c_2L}{2}\alpha_k^2 \gamma_k^2\\
		= &F(\theta_k)-\frac{c_1 \alpha_k\gamma_k}{2}\|\nabla F(\theta_k)\|^2 +\frac{c_1 \alpha_k\gamma_k}{2}\|b_k\|^2+\frac{c_2L}{2}\alpha_k^2 \gamma_k^2\\
		\overset{(c)}{\leq} &F(\theta_k)-\frac{c_1 \alpha_k\gamma_k}{2}\|\nabla F(\theta_k)\|^2 +\frac{c_1 c_3^2}{2}\alpha_k\gamma_k^3+\frac{c_2L}{2}\alpha_k^2 \gamma_k^2\\
	\end{split}
\end{equation}
where $(a)$ is by Lemmas \ref{biased_estimators} and \ref{norm}.
$(b)$ is due to $-\langle a,b\rangle\leq \frac{1}{2}\|a\|^2+\frac{1}{2}\|b\|^2$. $(c)$ is by Lemma \ref{bias-norm-lemma}.

By considering the telescoping sum, we get 
\begin{equation}
	\begin{split}
		\mathbb{E}[F(\theta_{K+1})|\mathcal{H}_K]
		&\leq F(\theta_0)-\frac{c_1}{2}\sum_{k=0}^{K}\alpha_k\gamma_k\|\nabla F(\theta_k)\|^2 \\&+\frac{c_1 c_3^2}{2}\sum_{k=0}^{K}\alpha_k\gamma_k^3+\frac{c_2L}{2}\sum_{k=0}^{K}\alpha_k^2\gamma_k^2\\
		0 \leq \mathbb{E}[\delta_{K+1}|\mathcal{H}_K]
		&\leq \delta_0-\frac{c_1}{2}\sum_{k=0}^{K}\alpha_k\gamma_k\|\nabla F(\theta_k)\|^2 \\&+\frac{c_1 c_3^2}{2}\sum_{k=0}^{K}\alpha_k\gamma_k^3+\frac{c_2L}{2}\sum_{k=0}^{K}\alpha_k^2\gamma_k^2\\
	\end{split}
\end{equation}
Hence, 
\begin{equation}\label{nablaF_2pt}
	\begin{split}
		&\sum_{k=0}^{K}\alpha_k\gamma_k\mathbb{E}[\|\nabla F(\theta_k)\|^2]\\
		\leq &\frac{2}{c_1}\mathbb{E}[\delta_0]+c_3^2\sum_{k=0}^{K}\alpha_k\gamma_k^3+\frac{c_2L}{c_1}\sum_{k=0}^{K}\alpha_k^2\gamma_k^2
	\end{split}
\end{equation}
The first term is bounded and by Assumption \ref{step_sizes_1},
\begin{equation}
	\lim_{K\rightarrow\infty}\sum_{k=0}^{K}\alpha_k\gamma_k^3<\infty\;\;\;\text{and}\;\;\;\lim_{K\rightarrow\infty}\sum_{k=0}^{K}\alpha_k^2\gamma_k^2<\infty.
\end{equation}  
We conclude that 
\begin{equation}\label{nablaF}
	\lim_{K\rightarrow\infty}\sum_{k=0}^{K}\alpha_k\gamma_k\mathbb{E}[\|\nabla F(\theta_k)\|^2]	<\infty.
\end{equation}
Moreover, since the series $\sum_k\alpha_k\gamma_k$ diverges by Assumption \ref{step_sizes_1}, we have
\begin{equation}
	\lim_{k\rightarrow\infty}\inf \mathbb{E}[\|\nabla F(\theta_k)\|^2]=0.
\end{equation}

To prove that $\lim_{k\rightarrow\infty}\mathbb{E}[\|\nabla F(\theta_k)\|^2]=0$, we consider the hypothesis:

(H) $\lim_{k\rightarrow\infty}\sup \mathbb{E}[\|\nabla F(\theta_k)\|^2]\geq \rho$ for an arbitrary $\rho>0$.

Assume (H) to be true. Then, we can always find an arbitrary subsequence $\big(\|\nabla F(\theta_{k_l})\|\big)_{l\in\mathbb{N}}$ of $\|\nabla F(\theta_k)\|$, such that $\|\nabla F(\theta_{k_l})\|\geq \rho - \varepsilon$, $\forall l$, for $\rho - \varepsilon>0$ and $\varepsilon>0$.

Then, by the $L$-smoothness property and applying the descent step of the algorithm, 
\begin{equation}
	\begin{split}
		&\|\nabla F(\theta_{k_l+1})\| \\ \geq &\|\nabla F(\theta_{k_l})\|-\|\nabla F(\theta_{k_l+1})-\nabla F(\theta_{k_l})\| \\
		\geq &\rho - \varepsilon-L\|\theta_{k_l+1}-\theta_{k_l}\|\\
		= &\rho - \varepsilon-L\alpha_{k_l}\|g_{k_l}\|.\\
	\end{split}
\end{equation}

Taking the expectation on both sides, we get
\begin{equation}
	\mathbb{E}[\|\nabla F(\theta_{k_l+1})\|]	\geq \rho - \varepsilon-L\sqrt{c_2}\alpha_{k_l}\gamma_{k_l},
\end{equation}
as by Jensen's inequality, we have $\big(\mathbb{E}[\|g_{k_l}\|]\big)^2\leq \mathbb{E}[\|g_{k_l}\|^2]\leq c_2\gamma_{k_l}^2$ by Lemma \ref{norm}, meaning $\mathbb{E}[\|g_{k_l}\|] \leq \sqrt{c_2}\gamma_{k_l}$ and finally $-\mathbb{E}[\|g_{k_l}\|] \geq -\sqrt{c_2}\gamma_{k_l}$.

Since $k_l\rightarrow\infty$ as $l\rightarrow\infty$, we can always find a subsequence of $(k_{l_p})_{p\in\mathbb{N}}$ such that $k_{l_{p+1}}-k_{l_p}>1$. As $\alpha_{k_l}\gamma_{k_l}$ is vanishing, we consider $(k_l)_{l\in\mathbb{N}}$ starting from $\alpha_{k_l}\gamma_{k_l}<\frac{\rho - \varepsilon}{L\sqrt{c_2}}$. Applying Jensen's inequality again,
\begin{equation}
	\begin{split}
		\mathbb{E}[\|\nabla F(\theta_{k_l+1})\|^2]&\geq \big(\mathbb{E}[\|\nabla F(\theta_{k_l+1})\|]\big)^2 \\ &\geq ( \rho - \varepsilon-L\sqrt{c_2}\alpha_{k_l}\gamma_{k_l})^2;
	\end{split}
\end{equation}
Thus, 
\begin{equation}
	\begin{split}
		&\sum_{k=0}^{\infty} \alpha_{k+1}\gamma_{k+1}\mathbb{E}[\|\nabla  F(\theta_{k+1})\|^2]\\ &\geq (\rho - \varepsilon)^2\sum_{k=0}^{\infty}\alpha_{k+1}\gamma_{k+1}-2(\rho - \varepsilon)L\sqrt{c_2}\sum_{k=0}^{\infty}\alpha_{k+1}\gamma_{k+1}\alpha_{k}\gamma_{k}\\&+ L^2c_2\sum_{k=0}^{\infty}\alpha_{k+1}\gamma_{k+1}\alpha_{k}^2\gamma_{k}^2\\
		&\geq (\rho - \varepsilon)^2\sum_{k=0}^{\infty}\alpha_{k+1}\gamma_{k+1}-2(\rho - \varepsilon)L\sqrt{c_2}\sum_{k=0}^{\infty}\alpha_{k}^2\gamma_{k}^2\\&+L^2c_2\sum_{k=0}^{\infty}\alpha_{k+1}\gamma_{k+1}\alpha_{k}^2\gamma_k^2\\
		&=+\infty,
	\end{split}
\end{equation}

as the first series diverges, and the second and the third converge by Assumption \ref{step_sizes_1}.
This implies that the series $\sum_{k} \alpha_{k}\gamma_{k}\mathbb{E}[\|\nabla  F(\theta_{k})\|^2]$ diverges. This is a contradiction as this series converges by (\ref{nablaF}). Therefore, hypothesis (H) cannot be true and $\mathbb{E}[\|\nabla F(\theta_{k})\|^2]$ converges to zero.

\subsection{Proof of Theorem \ref{th-ncvx-rate}: Convergence rate}\label{th-ncvx-rate-proof}
Considering again the $L$-smoothness inequality, we have
\begin{equation}
	\begin{split}
		F(\theta_{k+1})
		&\leq F(\theta_k)-\alpha_k\langle\nabla F(\theta_k), g_k\rangle +\frac{\alpha_k^2 L}{2}\|g_k\|^2.\\
	\end{split}
\end{equation}
Taking the conditional expectation given $\mathcal{H}_k$,
\begin{equation}\label{ncvx-rate}
	\begin{split}
		&F(\theta_{k+1}) \\
		\leq &F(\theta_k)-c_1\alpha_k\gamma_k\langle\nabla F(\theta_k), \nabla F(\theta_k)+b_k\rangle +\frac{c_2L}{2}\alpha_k^2 \gamma_k^2\\
		= &F(\theta_k)-c_1 \alpha_k\gamma_k\|\nabla F(\theta_k)\|^2-c_1 \alpha_k\gamma_k\langle\nabla F(\theta_k), b_k\rangle\\&+\frac{c_2L}{2}\alpha_k^2 \gamma_k^2\\
		\overset{(a)}{\leq} &F(\theta_k)-c_1 \alpha_k\gamma_k\|\nabla F(\theta_k)\|^2+\frac{c_1 \alpha_k\gamma_k}{2}\|\nabla F(\theta_k)\|^2 \\&+\frac{c_1 \alpha_k\gamma_k}{2}\|b_k\|^2+\frac{c_2L}{2}\alpha_k^2 \gamma_k^2\\
		= &F(\theta_k)-\frac{c_1 \alpha_k\gamma_k}{2}\|\nabla F(\theta_k)\|^2 +\frac{c_1 \alpha_k\gamma_k}{2}\|b_k\|^2+\frac{c_2L}{2}\alpha_k^2 \gamma_k^2\\
	\end{split}
\end{equation}
where $(a)$ is by $-\langle a,b\rangle\leq \frac{1}{2}\|a\|^2+\frac{1}{2}\|b\|^2$

Taking the telescoping sum of (\ref{ncvx-rate}), 

\begin{equation}
	\begin{split}
		\mathbb{E}[F(\theta_{K+1})|\mathcal{H}_K]
		&\leq F(\theta_0)-\frac{c_1}{2}\sum_{k}\alpha_k\gamma_k\|\nabla F(\theta_k)\|^2 \\&+\frac{c_1 }{2}\sum_{k}\alpha_k\gamma_k\|b_k\|^2+\frac{c_2L}{2}\sum_{k}\alpha_k^2\gamma_k^2\\
		0 \leq \mathbb{E}[\delta_{K+1}|\mathcal{H}_K]
		&\leq \delta_0-\frac{c_1}{2}\sum_{k}\alpha_k\gamma_k\|\nabla F(\theta_k)\|^2 \\&+\frac{c_1 }{2}\sum_{k}\alpha_k\gamma_k\|b_k\|^2+\frac{c_2L}{2}\sum_{k}\alpha_k^2\gamma_k^2\\
	\end{split}
\end{equation}
Hence, 
\begin{equation}\label{nablaF_2}
	\begin{split}
		&\sum_{k}\alpha_k\gamma_k\mathbb{E}[\|\nabla F(\theta_k)\|^2]\\	
		\leq &\frac{2}{c_1}\delta_0 +\sum_{k}\alpha_k\gamma_k\|b_k\|^2+\frac{ c_2 L}{c_1}\sum_{k}\alpha_k^2\gamma_k^2\\
		\leq &\frac{2}{c_1}\delta_0 +c_3^2\sum_{k}\alpha_k\gamma_k^3+\frac{ c_2 L}{c_1}\sum_{k}\alpha_k^2\gamma_k^2\\
	\end{split}
\end{equation}

Let $\alpha_k = \alpha_0(1+k)^{-\upsilon_1}$ and $\gamma_k = \gamma_0 (1+k)^{-\upsilon_2}$.
Then, to satisfy Assumption \ref{step_sizes_1}, it is sufficient to find $\upsilon_1$ and $\upsilon_2$ such that $0<\upsilon_1+\upsilon_2\leq 1$, $\upsilon_1+3\upsilon_2>1$, and $\upsilon_1+\upsilon_2>0.5$.

We know that, $\forall K>0$, 
\begin{equation}\label{a_0g_0}
	\begin{split}
		\sum_{k=0}^{K}\alpha_k\gamma_k^3 &= \alpha_0 \gamma_0^3 +\sum_{k=1}^{K}\alpha_k\gamma_k^3\\ &\leq \alpha_0 \gamma_0^3\bigg(1 + \int_{0}^{K}(x+1)^{-\upsilon_1-3\upsilon_2}dx\bigg)\\
		&= \alpha_0 \gamma_0^3\bigg(1 + \frac{1}{\upsilon_1+3\upsilon_2-1}-\frac{(K+1)^{-\upsilon_1-3\upsilon_2+1}}{\upsilon_1+3\upsilon_2-1}\bigg)\\
		&\leq \alpha_0 \gamma_0^3\bigg(1 + \frac{1}{\upsilon_1+3\upsilon_2-1}\bigg)\\
		&= \alpha_0 \gamma_0^3\bigg(\frac{\upsilon_1+3\upsilon_2}{\upsilon_1+3\upsilon_2-1}\bigg).\\\
	\end{split}
\end{equation}	
Similarly,
\begin{equation}
	\sum_{k=0}^{K}\alpha_k^2\gamma_k^2 \leq \alpha_0^2\gamma_0^2 \bigg(\frac{2\upsilon_1+2\upsilon_2}{2\upsilon_1+2\upsilon_2-1}\bigg)
\end{equation}
\begin{itemize}
	\item Next, when $0<\upsilon_1+\upsilon_2<1$,
	\begin{equation}
		\begin{split}
			\sum_{k=0}^{K}\alpha_k\gamma_k &\geq \alpha_0\gamma_0\int_{0}^{K+1} (x+1)^{-\upsilon_1-\upsilon_2} dx\\
			&= \frac{\alpha_0\gamma_0}{(1-\upsilon_1-\upsilon_2)} \bigg((K+2)^{1-\upsilon_1-\upsilon_2}-1\bigg).
		\end{split}
	\end{equation}
	Thus, making use of inequality (\ref{nablaF_2})
	\begin{equation}
		\begin{split}
			&\frac{\sum_{k}\alpha_k\gamma_k\mathbb{E}[\|\nabla F(\theta_k)\|^2]}{\sum_k \alpha_k\gamma_k} \\	
			\leq &\frac{(1-\upsilon_1-\upsilon_2)}{(K+2)^{1-\upsilon_1-\upsilon_2}-1}\times\\
			&\bigg[\frac{2\delta_0}{c_1\alpha_0\gamma_0}+ \frac{(\upsilon_1+3\upsilon_2)(c_3\gamma_0)^2}{\upsilon_1+3\upsilon_2-1}+\frac{2(\upsilon_1+\upsilon_2)c_2\alpha_0\gamma_0L}{c_1(2\upsilon_1+2\upsilon_2-1)}\bigg]\\
		\end{split}
	\end{equation}
	In the pursuit of optimizing the time-varying component, which follows the scaling of $O\left(\frac{1}{K^{1-\upsilon_1-\upsilon_2}}\right)$, we find that the most suitable values for the exponents are $\upsilon_1 = \upsilon_2 = \frac14$, resulting in a rate of $O\left(\frac{1}{\sqrt{K}}\right)$. However, it is worth noting that with this specific selection, the constant portion becomes excessively large, underscoring the need for a compromise.
	\item Otherwise, when $\upsilon_1+\upsilon_2=1$,
	\begin{equation}
		\begin{split}
			\sum_{k=0}^{K}\alpha_k\gamma_k &\geq \alpha_0\gamma_0\int_{0}^{K+1} \frac{1}{x+1} dx\\
			&= \alpha_0\gamma_0 \ln(K+2).
		\end{split}
	\end{equation}
	Thus, we get
	\begin{equation}
		\begin{split}
			&\frac{\sum_{k}\alpha_k\gamma_k\mathbb{E}[\|\nabla F(\theta_k)\|^2]}{\sum_k \alpha_k\gamma_k}\\	
			\leq &\frac{1}{\ln(K+2)}\times\\
			&\bigg[\frac{2\delta_0}{c_1\alpha_0\gamma_0}+ \frac{(\upsilon_1+3\upsilon_2)(c_3\gamma_0)^2}{\upsilon_1+3\upsilon_2-1}+\frac{2(\upsilon_1+\upsilon_2)c_2\alpha_0\gamma_0L}{c_1(2\upsilon_1+2\upsilon_2-1)}\bigg]\\
		\end{split}
	\end{equation}
\end{itemize}

\bibliography{example_paper}

\begin{thebibliography}{10}
\providecommand{\url}[1]{#1}
\csname url@samestyle\endcsname
\providecommand{\newblock}{\relax}
\providecommand{\bibinfo}[2]{#2}
\providecommand{\BIBentrySTDinterwordspacing}{\spaceskip=0pt\relax}
\providecommand{\BIBentryALTinterwordstretchfactor}{4}
\providecommand{\BIBentryALTinterwordspacing}{\spaceskip=\fontdimen2\font plus
\BIBentryALTinterwordstretchfactor\fontdimen3\font minus
  \fontdimen4\font\relax}
\providecommand{\BIBforeignlanguage}[2]{{%
\expandafter\ifx\csname l@#1\endcsname\relax
\typeout{** WARNING: IEEEtran.bst: No hyphenation pattern has been}%
\typeout{** loaded for the language `#1'. Using the pattern for}%
\typeout{** the default language instead.}%
\else
\language=\csname l@#1\endcsname
\fi
#2}}
\providecommand{\BIBdecl}{\relax}
\BIBdecl

\bibitem{FLGoogle}
\BIBentryALTinterwordspacing
K.~Bonawitz, H.~Eichner, W.~Grieskamp, D.~Huba, A.~Ingerman, V.~Ivanov,
  C.~Kiddon, J.~Kone\v{c}n\'{y}, S.~Mazzocchi, B.~McMahan, T.~Van~Overveldt,
  D.~Petrou, D.~Ramage, and J.~Roselander, ``Towards federated learning at
  scale: System design,'' in \emph{Proceedings of Machine Learning and
  Systems}, A.~Talwalkar, V.~Smith, and M.~Zaharia, Eds., vol.~1, 2019, pp.
  374--388. [Online]. Available:
  \url{https://proceedings.mlsys.org/paper_files/paper/2019/file/bd686fd640be98efaae0091fa301e613-Paper.pdf}
\BIBentrySTDinterwordspacing

\bibitem{FL1}
\BIBentryALTinterwordspacing
B.~McMahan, E.~Moore, D.~Ramage, S.~Hampson, and B.~A.~y. Arcas,
  ``{Communication-Efficient Learning of Deep Networks from Decentralized
  Data},'' in \emph{Proceedings of the 20th International Conference on
  Artificial Intelligence and Statistics}, ser. Proceedings of Machine Learning
  Research, A.~Singh and J.~Zhu, Eds., vol.~54.\hskip 1em plus 0.5em minus
  0.4em\relax PMLR, 20--22 Apr 2017, pp. 1273--1282. [Online]. Available:
  \url{https://proceedings.mlr.press/v54/mcmahan17a.html}
\BIBentrySTDinterwordspacing

\bibitem{FL2}
X.~Zhang, M.~Hong, S.~Dhople, W.~Yin, and Y.~Liu, ``Fedpd: A federated learning
  framework with adaptivity to non-iid data,'' \emph{IEEE Transactions on
  Signal Processing}, vol.~69, pp. 6055--6070, 2021.

\bibitem{FL3}
J.~Wang, Q.~Liu, H.~Liang, G.~Joshi, and H.~V. Poor, ``A novel framework for
  the analysis and design of heterogeneous federated learning,'' \emph{IEEE
  Transactions on Signal Processing}, vol.~69, pp. 5234--5249, 2021.

\bibitem{FLS}
\BIBentryALTinterwordspacing
A.~Elgabli, C.~B. Issaid, A.~S. Bedi, K.~Rajawat, M.~Bennis, and V.~Aggarwal,
  ``{F}ed{N}ew: A communication-efficient and privacy-preserving {N}ewton-type
  method for federated learning,'' in \emph{Proceedings of the 39th
  International Conference on Machine Learning}, ser. Proceedings of Machine
  Learning Research, K.~Chaudhuri, S.~Jegelka, L.~Song, C.~Szepesvari, G.~Niu,
  and S.~Sabato, Eds., vol. 162.\hskip 1em plus 0.5em minus 0.4em\relax PMLR,
  17--23 Jul 2022, pp. 5861--5877. [Online]. Available:
  \url{https://proceedings.mlr.press/v162/elgabli22a.html}
\BIBentrySTDinterwordspacing

\bibitem{FLS2}
T.~Li, A.~K. Sahu, M.~Zaheer, M.~Sanjabi, A.~Talwalkar, and V.~Smithy,
  ``Feddane: A federated newton-type method,'' in \emph{2019 53rd Asilomar
  Conference on Signals, Systems, and Computers}, 2019, pp. 1227--1231.

\bibitem{FLchallenges}
T.~Li, A.~K. Sahu, A.~Talwalkar, and V.~Smith, ``Federated learning:
  Challenges, methods, and future directions,'' \emph{IEEE Signal Processing
  Magazine}, vol.~37, no.~3, pp. 50--60, 2020.

\bibitem{localSGD}
\BIBentryALTinterwordspacing
A.~Khaled, K.~Mishchenko, and P.~Richtarik, ``Tighter theory for local sgd on
  identical and heterogeneous data,'' in \emph{Proceedings of the Twenty Third
  International Conference on Artificial Intelligence and Statistics}, ser.
  Proceedings of Machine Learning Research, S.~Chiappa and R.~Calandra, Eds.,
  vol. 108.\hskip 1em plus 0.5em minus 0.4em\relax PMLR, 26--28 Aug 2020, pp.
  4519--4529. [Online]. Available:
  \url{https://proceedings.mlr.press/v108/bayoumi20a.html}
\BIBentrySTDinterwordspacing

\bibitem{pp}
\BIBentryALTinterwordspacing
T.~Chen, G.~Giannakis, T.~Sun, and W.~Yin, ``Lag: Lazily aggregated gradient
  for communication-efficient distributed learning,'' in \emph{Advances in
  Neural Information Processing Systems}, S.~Bengio, H.~Wallach, H.~Larochelle,
  K.~Grauman, N.~Cesa-Bianchi, and R.~Garnett, Eds., vol.~31.\hskip 1em plus
  0.5em minus 0.4em\relax Curran Associates, Inc., 2018. [Online]. Available:
  \url{https://proceedings.neurips.cc/paper/2018/file/feecee9f1643651799ede2740927317a-Paper.pdf}
\BIBentrySTDinterwordspacing

\bibitem{CQ1}
\BIBentryALTinterwordspacing
J.~Konečný, H.~B. McMahan, F.~X. Yu, P.~Richtárik, A.~T. Suresh, and
  D.~Bacon, ``Federated learning: Strategies for improving communication
  efficiency,'' 2016. [Online]. Available:
  \url{https://arxiv.org/abs/1610.05492}
\BIBentrySTDinterwordspacing

\bibitem{CQ2}
\BIBentryALTinterwordspacing
S.~Khirirat, H.~R. Feyzmahdavian, and M.~Johansson, ``Distributed learning with
  compressed gradients,'' 2018. [Online]. Available:
  \url{https://arxiv.org/abs/1806.06573}
\BIBentrySTDinterwordspacing

\bibitem{CQ3}
A.~Elgabli, J.~Park, A.~S. Bedi, M.~Bennis, and V.~Aggarwal, ``Q-gadmm:
  Quantized group admm for communication efficient decentralized machine
  learning,'' in \emph{ICASSP 2020 - 2020 IEEE International Conference on
  Acoustics, Speech and Signal Processing (ICASSP)}, 2020, pp. 8876--8880.

\bibitem{CQ4}
\BIBentryALTinterwordspacing
K.~Mishchenko, E.~Gorbunov, M.~Takáč, and P.~Richtárik, ``Distributed
  learning with compressed gradient differences,'' 2019. [Online]. Available:
  \url{https://arxiv.org/abs/1901.09269}
\BIBentrySTDinterwordspacing

\bibitem{S1}
Y.~Chen, R.~S. Blum, M.~Takáč, and B.~M. Sadler, ``Distributed learning with
  sparsified gradient differences,'' \emph{IEEE Journal of Selected Topics in
  Signal Processing}, vol.~16, no.~3, pp. 585--600, 2022.

\bibitem{ref2}
J.~C. Duchi, M.~I. Jordan, M.~J. Wainwright, and A.~Wibisono, ``Optimal rates
  for zero-order convex optimization: The power of two function evaluations,''
  \emph{IEEE Transactions on Information Theory}, vol.~61, no.~5, pp.
  2788--2806, 2015.

\bibitem{ref2-ref}
A.~Agarwal, O.~Dekel, and L.~Xiao, ``Optimal algorithms for online convex
  optimization with multi-point bandit feedback,'' in \emph{COLT}, 2010.

\bibitem{allerton}
\BIBentryALTinterwordspacing
E.~Mhanna and M.~Assaad, ``Rendering wireless environments useful for gradient
  estimators: A zero-order stochastic federated learning method,'' 2024.
  [Online]. Available: \url{https://arxiv.org/abs/2401.17460}
\BIBentrySTDinterwordspacing

\bibitem{ZO-Bay}
\BIBentryALTinterwordspacing
Z.~Dai, B.~K.~H. Low, and P.~Jaillet, ``Federated bayesian optimization via
  thompson sampling,'' in \emph{Advances in Neural Information Processing
  Systems}, H.~Larochelle, M.~Ranzato, R.~Hadsell, M.~Balcan, and H.~Lin, Eds.,
  vol.~33.\hskip 1em plus 0.5em minus 0.4em\relax Curran Associates, Inc.,
  2020, pp. 9687--9699. [Online]. Available:
  \url{https://proceedings.neurips.cc/paper_files/paper/2020/file/6dfe08eda761bd321f8a9b239f6f4ec3-Paper.pdf}
\BIBentrySTDinterwordspacing

\bibitem{Ereference}
B.~Moons, D.~Bankman, and M.~Verhelst, \emph{Embedded Deep Learning:
  Algorithms, Architectures and Circuits for Always-on Neural Network
  Processing}.\hskip 1em plus 0.5em minus 0.4em\relax Cham, Switzerland:
  Springer, 2018.

\bibitem{Kim}
M.~Kim, W.~Saad, M.~Mozaffari, and M.~Debbah, ``Green, quantized federated
  learning over wireless networks: An energy-efficient design,'' \emph{IEEE
  Transactions on Wireless Communications}, vol.~23, no.~2, pp. 1386--1402,
  2024.

\end{thebibliography}
\bibliographystyle{IEEEtran}

\end{document}